\documentclass[conference]{IEEEtran}
\IEEEoverridecommandlockouts
\usepackage{cite}
\usepackage{amsmath,amssymb,amsfonts}
\usepackage{algorithm}
\usepackage{algpseudocode}
\usepackage{multirow}
\usepackage{graphicx}
\usepackage{textcomp}
 \usepackage{hyperref}
\usepackage{xcolor}
\def\BibTeX{{\rm B\kern-.05em{\sc i\kern-.025em b}\kern-.08em
    T\kern-.1667em\lower.7ex\hbox{E}\kern-.125emX}}
\IEEEaftertitletext{%
  \begin{center}
  \begin{minipage}{0.96\textwidth}
    \footnotesize
    \textcopyright\ 2026, copyrighted by International Institute of Applied Informatics.\\
    DOI:
    \href{https://doi.org/10.23919/IIAI-AAICPS00095.2026.00072}
         {10.23919/IIAI-AAICPS00095.2026.00072}
  \end{minipage}
  \end{center}
}
\begin{document}

\title{Continual Learning for\\Reliable In-Sequence Molecular Communication}

\author{
\IEEEauthorblockN{Siddhant Setia}
\IEEEauthorblockA{\textit{Department of Computer Science} \\
\textit{University of Massachusetts, Boston}\\
Boston, MA, 02125, USA \\
siddhant.setia001@umb.edu}
\and
\IEEEauthorblockN{Junichi Suzuki}
\IEEEauthorblockA{\textit{Department of Computer Science} \\
\textit{University of Massachusetts, Boston}\\
Boston, MA, 02125, USA \\
jxs@cs.umb.edu}
\and
\IEEEauthorblockN{Tadashi Nakano}
\IEEEauthorblockA{\textit{Graduate School of Informatics} \\
\textit{Osaka Metropolitan University}\\
Osaka, Japan \\
tnakano@omu.ac.jp}
}

\maketitle

\begin{abstract}

This paper considers reliable in-sequence molecular communication, where bio-nanomachines exchange molecule-encoded information in a specified order, and investigates a new simulation-based method to incrementally evaluate communication protocols by gradually increasing the complexity of experimental settings. The proposed method, called Affine Memory Replay (AMR), builds a continual learning (CL) model to estimate communication performance between bio-nanomachines. AMR stores past simulation samples and synthesizes additional replay samples through interpolation and extrapolation between past samples. Through incremental performance estimation tasks, it replays both past and synthetic samples to regularize model updates, helping the model adapt to new experimental settings while retaining knowledge from earlier ones. Experimental results show that AMR significantly outperforms existing CL methods in maintaining overall estimation accuracy across sequentially learned estimation tasks. It effectively balances memory stability and learning plasticity under moderate training cost. 

\end{abstract}

\section{Introduction}
Molecular communication is a communication paradigm for networking {\it bio-nanomachines} such as modified biological cells and artificial cells (cell-like artificial nanostructures engineered with biological materials)~\cite{suda05exploratory,nakano13molcom,suzuki17molcom}. Using molecules as information carriers among bio-nanomachines, it is expected to enable various biomedical and healthcare applications, including in-body physiological sensing, biomedical anomaly detection, targeted drug delivery,  cellular/molecular-level medical procedures, and artificial morphogenesis~\cite{chahibi17drug,almpanis16survey,xiao24mc}.  

Molecular communication research often advances stepwise. When a communication protocol is studied, for example, it may first be examined under the simplest setting, such as a 1D noiseless space, to evaluate its characteristics and effectiveness. Then, it can be evaluated under more realistic settings by gradually introducing additional requirements or relaxing prior assumptions. The protocol may be examined in 2D and then 3D spaces by considering noise. While the transmitter~(Tx) and receiver~(Rx) bio-nanomachines may initially be placed near each other, the Tx-Rx distance can be gradually expanded, potentially with relay bio-nanomachines. Bio-nanomachines may eventually become mobile nodes. 

As the complexity of these experimental settings increases, simulation studies are conducted in addition to analytical work. When these simulations are computationally expensive, they are often augmented by machine learning (ML) methods. For example, if an ML model is built for a protocol, it can be trained on prior simulation results to estimate  performance measures such as latency and error rate.
This can reduce the computational cost of running additional simulations. 

Since protocol studies often proceed stepwise, simulation results are collected sequentially across increasingly complex settings. The ML model must therefore be updated as new simulation results become available, preferably without being retrained from scratch. For example, when transitioning from 1D to 2D spaces, a model previously trained on 1D simulation results is updated with 2D results. However, the model ``forgets'' the 1D setting and yields poor performance estimates for that setting, although it works well in 2D spaces. If the model is trained further with 3D results, it forgets the 1D setting even more and becomes even less capable of estimating performance there. This is called ``forgetting'' in the ML research~\cite{mccloskey89catastrophic}. The challenge is to have an ML model acquire new knowledge from sequentially collected simulation results, integrate it with current knowledge, and retain previously-learned knowledge effectively.

To address this challenge, this paper proposes a continual learning (CL) method that adapts incrementally to new experimental settings for communication latency estimation while retaining knowledge from earlier estimation tasks. It is evaluated with a feedback-based protocol for bio-nanomachines to perform in-sequence and timing-controlled  communication, in which they transmit and receive a series of information-carrying molecules ({\it information molecules}) in a certain order. The protocol employs Stop-and-Wait Automatic Repeat Request (SW-ARQ)~\cite{burton72error}, which is feasible to realize with simple computation and memory functions in bio-nanomachines. 

The proposed CL method, Affine Memory Replay~(AMR), operates on a simple feedforward neural network (NN). It synthesizes new training samples from past training samples (i.e., prior simulation results), buffer both past and synthetic samples, and replays them during later training stages to reinforce past knowledge. Replay performance is used to regularize updates to the NN model parameters (e.g., weights and biases). This replay strategy allows AMR to retain knowledge from earlier experimental settings while adapting to newly introduced settings, thereby balancing memory stability and learning plasticity in sequential performance estimation.

Experimental results demonstrate that AMR maintains higher overall RTT estimation accuracy than existing CL methods across sequentially learned estimation tasks by reducing forgetting while preserving competitive learning plasticity under moderate training cost.

\begin{figure*}[t]
  \centering
	\resizebox{1.0\linewidth}{!}{\includegraphics{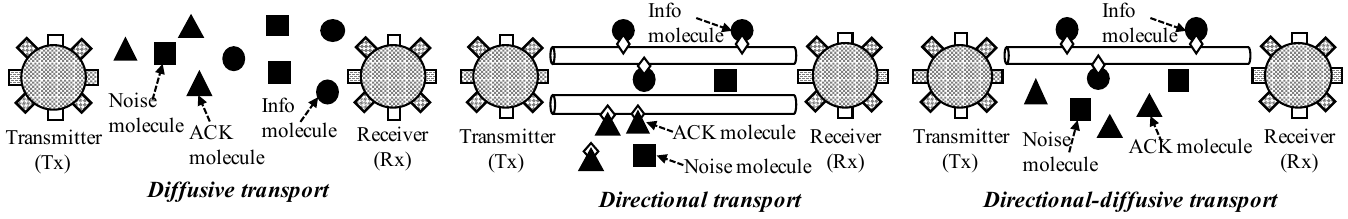}}
	\vspace{-1.75\baselineskip}
	\caption{Diffusive, Directional and Hybrid Transports with SW-ARQ in Molecular Communication}
	\label{fig:transports}
\end{figure*}

\vspace{-0.25\baselineskip}
\begin{figure}[t]
  \centering
	\resizebox{1.0\linewidth}{!}{\includegraphics{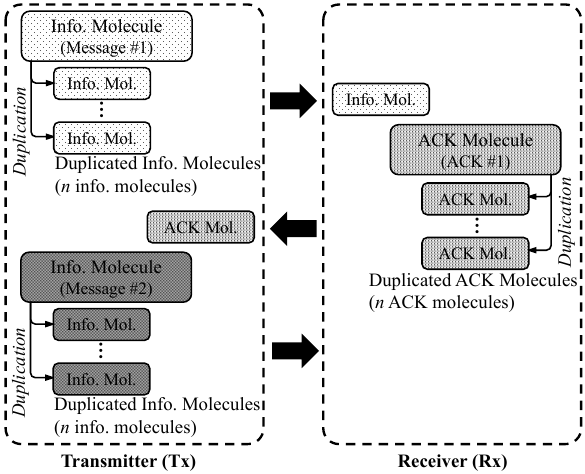}}
	\vspace{-1.75\baselineskip}
	\caption{Interaction between the Tx and the Rx}
	\label{fig:txrx}
\end{figure}

\section{Background and Related Work}

In-sequence molecule delivery can play a critical role in various 
applications. In artificial morphogenesis applications in regenerative medicine, bio-nanomachines made of living cells are designed to divide and grow into 3D multi-cellular structures such as tissues and organs. Molecular communication allows them to interact with each other using {\it artificial morphogens (AMs)}, which are modeled as  information molecules that encode morphological information. Reliable in-sequence molecule delivery provides a foundation for guiding the growth and differentiation of bio-nanomachines by controlling the order and timing of propagated AMs. 

In drug delivery applications, therapeutic bio-nanomachines leverage natural molecular signals in the body, 
or molecular signals released by other bio-nanomachines, to identify target sites and release drug molecules, thereby reducing side-effects at non-target sites. Reliable in-sequence delivery allows bio-nanomachines to modulate tumor behavior by controlling the order and duration of drug molecule exposure. 

To enhance the reliability of molecular communication, several feedback-based protocols have been studied.  
Feedback-based rate control schemes are proposed 
in~\cite{nakano13jsac,felicetti14tcpmol}.
In-sequence and at-least-once molecule delivery schemes are examined with SW-ARQ protocols in~\cite{wang14arq,bai15arq-bacteria,singh23ecc,mitzman15swarq}.  
None of these protocols use ML; this paper extends this line of work by developing a CL method for performance estimation.

Various ML methods have been studied for molecular communication; e.g.~\cite{cheng2025bio,cheng25dl,lee2017mlchannel,yilmaz2017ml,casaleiro2024sync,cheng2025localtrack,kara2022idx}. Most of them perform regression tasks, as in this paper, while~\cite{kara2022idx} considers a classification task, using a convolutional NN (CNN) to predict the source of each information molecule for molecular index modulation. Cheng et al. estimate the arrival time of propagated molecules at the Rx~\cite{cheng2025bio,cheng25dl}.
They use a transformer for a 2D noiseless environment in~\cite{cheng2025bio} and a multi-layer perceptron for a 3D noiseless environment in~\cite{cheng25dl}.
In~\cite{lee2017mlchannel,yilmaz2017ml}, NNs are used to estimate the fraction of molecules that arrive at the Rx by a certain deadline in 3D noiseless environments. In~\cite{casaleiro2024sync}, a CNN and a recurrent NN (RNN) are used to predict the type of information encoded in a molecule in a 3D noisy environment. In~\cite{cheng2025localtrack}, the location of a mobile Tx is estimated with a transformer in a 2D noiseless environment. None of these ML-based studies employ CL; their ML models require retraining from scratch and may suffer from forgetting if they are incrementally trained on different experimental settings.

This paper builds on the authors' prior work, which studied the impacts of CL in molecular communication by customizing existing methods~\cite{setia25bict}. In contrast, the present study develops a new CL method and evaluates it against existing methods. 

CL methods can be classified into three approaches: regularization-, replay- and architecture-based~\cite{DeLange22survey,clsurvey2024}. Regularization constrains updates to NN model parameters (e.g., weights and biases), thereby reducing interference with previously learned knowledge. Replay reintroduces information from previous tasks during training to reinforce past knowledge.
There are three major replay strategies: (1) experience replay (or rehearsal), which reuses past training samples; (2) generative replay, which synthesizes new samples from past data with a dedicated generator;  
and (3) feature/latent replay, which replays features (or embeddings) rather than raw samples. Architecture-based methods modify the NN model structure, for example, by 
expanding network capacity. This paper adopts a hybrid rehearsal-regularization approach for AMR: it replays past and synthetic samples and uses replay loss to regularize training and discourage parameter updates that degrade performance on replayed samples.

\section{Molecular Communication Model}
\label{sec:molcom}

This paper assumes a bounded 3D aqueous medium where the transmitter (Tx) and receiver (Rx) bio-nanomachines exchange information and acknowledgement (ACK) molecules.
Motivated by wet-lab experiments of molecular communication
(e.g.,~\cite{ortiz12}), DNA molecules are used as information and ACK molecules that can encode data as nucleotide sequences.

This paper considers three types of communication transports: 
(1) {\it diffusive transports}, 
(2) {\it directional transports} where molecules move directionally move on pre-defined protein filaments (e.g., microtubules and actin filaments) with molecular motors, 
and (3) {\it diffusive-directional hybrid transports} that combines diffusive and directional transports (Fig.~\ref{fig:transports}).
The Tx releases information molecules, each of which encodes a certain
message and carries it to the Rx (Fig.~\ref{fig:txrx}). The Rx is assumed to capture an
information molecule when the molecule physically contacts with the Rx. 
The Rx releases ACK molecules, each of which encodes
an acknowledgement of the message. The Tx is assumed to capture an ACK molecule when it contacts the Tx's surface. Noise molecules are stationary and interfere (i.e., collide) with other molecules.  

In a diffusive transport, a molecule diffuses via random thermal motion, which is governed by the diffusion coefficient~$D$ in each dimension: $D = \partial x^2 / (2 \times \partial t)$ where $x$ denotes the distance of molecular movement over time $t$ in a particular dimension. When a molecule collides with another molecule, it randomly moves to another position according to $D$.  

A directional transport utilizes a microtubule between the Tx and Rx. Based on findings in biomedical engineering (e.g.,~\cite{vale96nature}), the microtubule is configured to guide molecular motors (e.g., kinesin, dynein, and myosin) directionally. Molecules are attached to molecular motors that travel at a constant velocity for an expected distance before diffusing away. Collision with another molecule also causes diffusion. While diffusing, a molecule with a molecular motor may contact the microtubule and resume directional transport.

A hybrid transport uses both diffusive and directional transports. This paper deploys a directional transport for information molecules and a diffusive transport for ACK molecules.

Promising approaches to engineer bio-nanomachines with communication
capability include modifying biological cells and producing artificial cell-like structures using biological materials~\cite{nakano13molcom-nanobio}. 
Here, bio-nanomachines are modeled as modified biological cells that can possess communication-related functions, including transmission and reception functions, logic gates, toggle switches (i.e., 1-bit memories) and oscillators (i.e., clocks). These functions allow them to synthesize, release and capture specific molecules, trigger programmed chemical responses, retain communication states and control the timing of molecule release.

This paper employs SW-ARQ for molecular transmissions. The Tx releases $n$ duplicated information molecules that encode a particular message (e.g., Message 1 in Figs.~\ref{fig:txrx}). Once the Rx receives one of them, it releases $n$ duplicated ACK molecules. It does not respond to subsequent arrivals of the information molecules that encode the same message. Upon receiving one of the ACK molecules, the Tx releases $n$ information molecules that encode the next message (e.g., Message 2 in Figs.~\ref{fig:txrx}). It does not respond to subsequent arrivals of the ACK molecules for the previous message. 

If the Tx does not receive an ACK molecule 
for a message within a certain interval, called retransmission timeout interval (RTO), it retransmits another set of $n$ information molecules for the message. Molecule retransmission may occur multiple times until the Tx receives an ACK molecule or exceeds a predefined number of retransmissions. If the Rx does not receive any information molecules that encode the $i$-th message within RTO after sending ACK molecules for the ($i-$1)-th message, it retransmits $n$ ACK molecules for the previous message.

\section{Continual Performance Estimation}

This paper uses the protocol described in Section~\ref{sec:molcom} as a testbed for CL-based performance estimation tasks. Each task is defined by experimental settings,  including environmental parameters (e.g.,  environment size, noise level and Tx-Rx distance) and communication parameters (e.g., transport choice, the number of duplicated information/ACK molecules and RTO). The goal is to estimate the roundtrip time (RTT) of message delivery under those settings. RTT is measured as the time from the Tx releasing information molecules for a message to receiving the corresponding ACK molecule.

The RTT estimation model takes experimental settings as input and outputs an estimated RTT value. The model is trained incrementally through estimation tasks. Each task $t$ runs a series of simulations under the current experimental settings and feeds simulation results $D_t$ to the model to perform a supervised training session. Therefore, given the total number of tasks $T$, the model is trained on a stream of simulation results
$\{\mathcal{D}_1, \mathcal{D}_2, \dots, \mathcal{D}_t, \dots , \mathcal{D}_T \}$. $D_t$ consists of training samples $(x, y)$, where $x$ is the input (experimental settings) and $y$ is the estimation target (RTT). 
The model's objective is to optimize the estimation function $f_\theta(x)$ for the current task and all 
prior tasks. Formally, this requires finding optimized model parameters $\theta^*$ (weights and biases in the underlying neural network) that minimize the cumulative training loss over all tasks observed so far:

\vspace{-0.25\baselineskip}
\begin{equation}
\label{eq:basic_cl_loss}
\theta^* = \arg\min_{\theta} \sum_{t=1}^{t_c} \mathcal{L}_t
\end{equation}
\vspace{-0.25\baselineskip}

\noindent where $\mathcal{L}_t$ is the loss for task $t$ and $t_c$ is the current task index. 

Algorithm~\ref{alg:amr} shows the replay-based training procedure of AMR. AMR maintains two bounded replay buffers: a primary buffer~$\mathcal{M}$ to store past samples and a secondary buffer~$\mathcal{M}_s$ to store synthetic samples. One sample is drawn from each buffer (Lines~4 and 5) to compute the loss $\mathcal{L}_t$ as follows:  

\vspace{-0.25\baselineskip}
\begin{equation}
\begin{split}
\label{eq:cmr_loss}
\mathcal{L}_t = & \mathbb{E}_{(x,y)\sim \mathcal{D}_t} \left[ \ell\big(f_{\theta}(x), y\big) \right] \\
& + \alpha \cdot \mathbb{E}_{(x',z')\sim \mathcal{M}} \left[ \ell\big(f_{\theta}(x'), z'\big) \right] \\
& + \beta \cdot \mathbb{E}_{(x'',z'')\sim \mathcal{M}_s} \left[ \ell\big(f_{\theta}(x''), z''\big) \right]
\end{split}
\end{equation}
\vspace{-0.25\baselineskip}

\noindent where $\ell$ denotes the mean squared error loss. The first, second and third terms correspond to the losses for the current-task sample $(x, y)$, replayed past sample $(x', z')$ and replayed synthetic sample $(x'', z'')$, respectively. The computed $\mathcal{L}_t $ is used to update $\theta$ (Lines~7 and~8 in Algorithm~\ref{alg:amr}). AMR therefore augments current-task training with replay losses from the primary and secondary buffers, thereby regularizing updates to $\theta$ with respect to replayed samples. This helps mitigate forgetting by discouraging updates that degrade replay performance. It can also promote backward knowledge transfer by allowing current-task updates to improve performance on previously learned tasks.
The hyperparameters $\alpha$ and $\beta$ control the relative influence of primary- and secondary-buffer replay.

After the parameter updates, AMR updates the primary buffer~$\mathcal{M}$ with the current input and its estimated target~$(x,z)$ (Line~9 in Algorithm~\ref{alg:amr}). This update uses reservoir sampling~\cite{metaER}, which inserts a new sample with probability $C/n$, where $C$ is the buffer capacity and $n$ is the number of samples observed so far. If selected, the new sample replaces a uniformly chosen sample stored in the buffer (Algorithm~\ref{alg:reservoir_sampling}). This procedure keeps the buffer bounded while maintaining a representative set of past samples.

\begin{algorithm}
\caption{Affine Memory Replay}
\label{alg:amr}
\begin{algorithmic}[1]
\Require prediction function $f_\theta$; model parameters $\theta$; training dataset $\mathcal{D}_t$ for task $t$; 
total number of tasks $T$; primary and secondary replay buffers $\mathcal{M}$, $\mathcal{M}_s$; hyperparameters $\alpha$, $\beta$; learning rate $\lambda$, synthesis range parameter $q$
\State $\mathcal{M}, \mathcal{M}_s \leftarrow \{\}$
\For {each $t = 1,2, ..., T$}
    \For{each $(x, y)$ in $\mathcal{D}_t$}
        \State $(x', z') \gets \text{random}(\mathcal{M})$
        \State $(x'', z'') \gets \text{random}(\mathcal{M}_s)$
        \State $z \gets f_\theta(x)$
        \State $\text{reg} \gets \alpha \, L_{mse}(z', f_\theta(x')) + \beta\, L_{mse}(z'', f_\theta(x''))$
        \State $\theta \gets \theta - \lambda \cdot \nabla_\theta [ L_{mse}(y, f_\theta(x)) + \text{reg} ]$
        \State $\mathcal{M} \gets \text{reservoirSampling}(\mathcal{M}, (x, z))$
        \State $r \gets \text{uniformSampling}(-q, 1+q)$
        \If{$ 0 \leq r \leq 1$}
            \State $\mathcal{M}_s \gets \text{interpolateReplaySample}(\mathcal{M}, \mathcal{M}_s, r)$
        \Else
           \State $\mathcal{M}_s \gets \text{extrapolateReplaySample}(\mathcal{M}, \mathcal{M}_s, r)$
        \EndIf
    \EndFor
\EndFor
\end{algorithmic}
\end{algorithm}

\vspace{-1.0\baselineskip}
\begin{algorithm}
\caption{reservoirSampling()}
\label{alg:reservoir_sampling}
\begin{algorithmic}[1]
\State \textbf{Input:}  replay buffer $\mathcal{M}$, incoming sample ($x$, $y$), buffer capacity $C$, total number of observed samples $n$
\If{$C > |\mathcal{M}|$}
    \State $\mathcal{M}[n] \leftarrow (x, y)$
\Else
    \State $j = \text{randomInt} (0, n-1)$
    \If{$j < |\mathcal{M}|$}
        \State $\mathcal{M}[j] \leftarrow (x, y)$
    \EndIf
\EndIf
\State $n \gets n + 1$
\State \textbf{return} $\mathcal{M}$
\end{algorithmic}
\end{algorithm}

AMR then synthesizes a new sample using two past samples drawn from the primary buffer~$\mathcal{M}$~(Lines~10 to 15 in Algorithm~\ref{alg:amr}). A random factor $r$ is sampled from a uniform distribution over $(-q, 1+q)$, where $q$ controls how far synthetic samples may extend beyond the line segment between the two past samples. If $0 \leq r \leq 1$, AMR uses interpolation; otherwise, it uses extrapolation.

For interpolation, AMR generates a synthetic input $x_s$ as a convex combination of two sampled inputs (Lines~1 to~3 in Algorithm~\ref{alg:convex_synthesis}). $x_s$ lies inside the convex hull of the two inputs (i.e., on the line segment between them). By transforming past samples into a convex neighborhood, AMR expands the coverage of the replay space and encourages a smoother, better-generalized approximation of previously learned tasks during later training stages. 
After constructing $x_s$, AMR computes its Euclidean distances to the two sampled inputs, $d_1$ and $d_2$ (Lines~4 and~5), and uses these distances to compute a synthetic target value $z_s$ through distance-weighted interpolation (Line~6). This way, the synthetic sample $(x_s, z_s)$ preserves information from past samples while introducing additional diversity into the replay process. 

If extrapolation is selected (Line~14 in  Algorithm~\ref{alg:amr}), $x_s$ is generated outside the line segment between the two sampled inputs, thereby extending the replay space beyond the convex hull of past samples  (Lines~1 to~3 in Algorithm~\ref{alg:affine_synthesis}). After constructing $x_s$, AMR computes the synthetic target value $z_s$ by performing linear extrapolation along the affine line defined by the two sampled pairs (Line~4).
This allows AMR to generate synthetic samples that preserve the directional trend between past samples while increasing the diversity and coverage of the replay space.

Finally, the generated synthetic sample is inserted into the secondary buffer~$\mathcal{M}_s$ using reservoir sampling.

\begin{algorithm}
\caption{interpolateReplaySample()}
\label{alg:convex_synthesis}
\begin{algorithmic}[1]
\Require primary and secondary replay buffers $\mathcal{M}$, $\mathcal{M}_s$; random factor $r$
\State $(x_c, z_c) \gets$ random($\mathcal{M}$)
\State $(x, z) \gets $ random($\mathcal{M}$)
\State $x_s \gets x_c + r \times (x - x_c)$
\State $d_1 \gets \text{dist}(x_s, x)$
\State $d_2 \gets \text{dist}(x_c, x_s)$
\State $z_s \gets \dfrac{d_2 \cdot z + d_1 \cdot z_c}{d_1 + d_2}$
\State $\textbf{return} \text{ } \text{reservoirSampling}(\mathcal{M}_s, (x_s, z_s))$
\end{algorithmic}
\end{algorithm}

\vspace{-1.0\baselineskip}
\begin{algorithm}
\caption{extrapolateReplaySample()}
\label{alg:affine_synthesis}
\begin{algorithmic}[1]
\Require primary and secondary replay buffers $\mathcal{M}$, $\mathcal{M}_s$; random factor $r$
\State $(x_c, z_c) \gets$ random($\mathcal{M}$)
\State $(x, z) \gets $ random($\mathcal{M} - x_c$)
\State $x_s \gets x_c + r \times (x - x_c)$
\State $z_s \gets z_c + r \times (z - z_c)$
\State $\textbf{return} \text{ } \text{reservoirSampling}(\mathcal{M}_s, (x_s, z_s))$
\end{algorithmic}
\end{algorithm}

\section{Experimental Results}

This section evaluates AMR and other CL methods, namely CLeaR~\cite{he21clear}, ACLT~\cite{solarRad_DER} and A2ER~\cite{a2er}, through Monte Carlo (particle-based) molecular communication simulations. 
 
This paper implements AMR on a two-layer feedforward neural network with a fully connected hidden layer with ReLU activation. Input and output values are normalized to [0, 1] with min-max normalization. 

Table~\ref{tab:train-hyper-params} shows the hyperparameter values for AMR and the other methods. Those for AMR and CLeaR are selected through grid search on a subset of the training dataset. The hyperparameters for ACLT and A2ER are adopted from the original papers~\cite{solarRad_DER,a2er}. 

Each method is trained using the stochastic gradient descent (SGD) optimizer for 100 epochs with a batch size of 1 and a learning rate of 0.001. The loss function is the mean squared error (MSE). Early stopping is used for all methods with a patience of 10 epochs and a minimum improvement threshold (min delta) of 0.003. 

\vspace{-1.0\baselineskip}
\begin{table}[h]
\centering
\setlength{\tabcolsep}{4pt}
\caption{Hyperparameter configuration}
\vspace{-0.5\baselineskip}
    \begin{tabular}{| c | c | c || c | c | c |}
    \hline
     Method & Parameter & Value & Method  & Parameter & Value \\
    \hline
    \hline
    \multirow{6}{*}{AMR}
     & $\alpha$ & 1   & \multirow{3}{*}{CLeaR}   & $\alpha$      & 0.5 \\
     & $\beta$  & 0.5 &                         & $\lambda$     & 2 \\
     & Primary buffer size & 50 &                  & Buffer size   & 50  \\
     & Secondary buffer size & 50 &             &               &     \\
     & Synthesis range param $q$ & 0.1 &             &               &     \\
    \hline
    \multirow{6}{*}{ACLT}
     & $\alpha$ & 0.2 & \multirow{7}{*}{A2ER} & $\alpha$      & 1   \\
     & $\beta$  & 0.2 &                        & $\beta$       & 0.5 \\
     & $\gamma$ & 0.2 &                        & Buffer Size   & 50  \\
     & Main buffer size & 50 &                  & $\rho$        & 0.5 \\
     & Soft buffer size & 1  &                  & $\epsilon$    & $10^{-12}$ \\
     &                  &    &                  & $\lambda$     & 0.5 \\
     &                  &    &                  & $\Delta Q$    & 0   \\
    \hline
    \end{tabular}
\label{tab:train-hyper-params}
\end{table}

\subsection{Simulation Settings and Evaluation Metrics}

Table~\ref{tab:param-mc} shows the simulation settings for molecular communication. Each training and test data sample is prepared with the median RTT obtained from 500 independent simulations.

The boundaries of the environment are simulated in a non-repulsive manner. Molecules do not rebound against boundaries. Molecule-to-molecule collisions are also simulated in a non-repulsive manner. Each molecule is prohibited to move to the location of another molecule. Colliding molecules do not rebound with each other.

\begin{table}[htb]
    \centering
    \caption {Simulation Settings for Molecular Communication}
    \vspace{-0.5\baselineskip}
    \begin{tabular}{|l||c|}
    \hline
    {\bf Parameter} & {\bf Value} \\
    \hline \hline 
    Size length of the environment & 150 $\mu$m \\
    \hline 
    Diameter of Tx and Rx & 5 $\mu$m \\
    \hline
    Tx to Rx distance & 10 to 90 $\mu$m \\
    \hline 
    Diameter of an information/ACK molecule & 1 $\mu$m \\
    \hline 
    Diameter of a noise molecule & 1 $\mu$m \\
    \hline
    Number of noise molecules & 0 to 10$^5$   \\
    \hline 
    Diffusion coefficient ($D$) & 0.5 \\
    \hline 
    Velocity of a molecular motor on a microtubule   & 1 $\mu$m/s \\
    \hline
    \begin{tabular}{@{}l@{}}Expected travel distance of\\a molecular motor on a microtubule\end{tabular} & 4 $\mu$m \\
    \hline
    \# of duplicated molecules per message/acknowledgement & 10 \\
    \hline
    Max number of molecule retransmissions & 5 \\
    \hline 
    \end{tabular}
\label{tab:param-mc}
\end{table}

\begin{figure}[h]
    \centering
    \resizebox{1.0\linewidth}{!}{\includegraphics{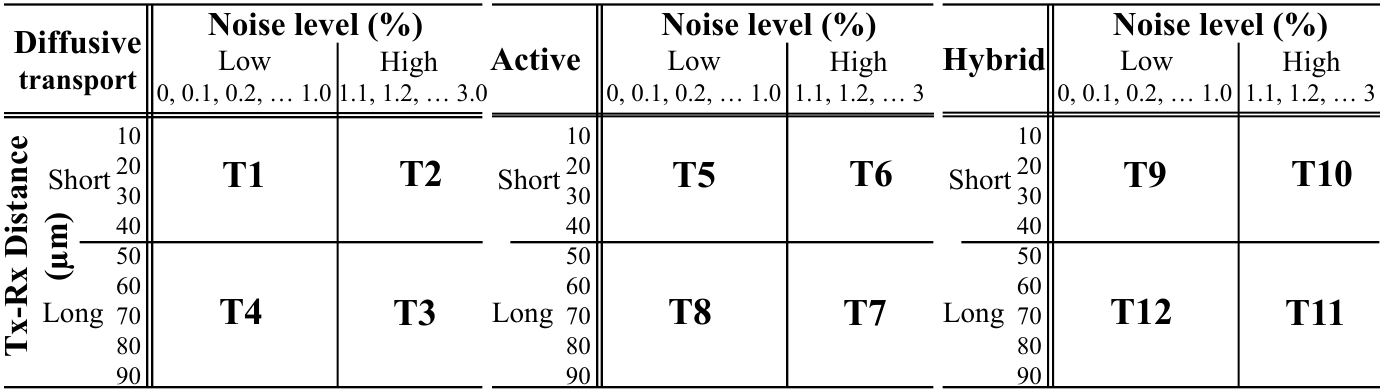}}
    \vspace{-1.5\baselineskip}
    \caption{RTT Estimation Tasks (T1 to T12)}
    \label{fig:tasks}
\end{figure}


Each CL method (RTT estimation model) is trained incrementally on 12 estimation tasks shown in Fig.~\ref{fig:tasks}. In the first task (T1), for example, simulations are carried out using diffusive transport for each pair of Tx-Rx distance and noise level to obtain RTT values. The Tx-Rx distance, noise level and other experimental settings are combined into a training input, and the median RTT is used as the estimation target. 

During training, each CL method has no access to training data from prior tasks. At test time, it performs RTT estimation for the current and all prior tasks. It does not know which task a given test input belongs to. For example, when a CL method is trained on the second task (T2), it cannot use training data from T1 but can use only T2 data. The method is then tested by estimating RTT for both T1 and T2. 

Estimation tasks are arranged into $T$ evaluation scenarios, where $T$ is the total number of tasks ($T$ = 12 in this paper). The $t$-th scenario evaluates each CL method after it has been sequentially trained on $t$ tasks. 
The method is evaluated using the mean squared error between the estimated and true RTT values ($\mathcal{L_\text{mse}}$). After all scenarios are completed, each method produces the following matrix of test errors. 

\vspace{-0.5\baselineskip}
\[
\begin{bmatrix}
  \mathcal{L_\text{mse}}_1^1 \\
  \mathcal{L_\text{mse}}_1^2 & \mathcal{L_\text{mse}}_2^2 \\
  \mathcal{L_\text{mse}}_1^3 & \mathcal{L_\text{mse}}_2^3 & \mathcal{L_\text{mse}}_3^3 \\
  \vdots & \vdots & \vdots & \ddots \\
  \mathcal{L_\text{mse}}_1^{T-1} & \mathcal{L_\text{mse}}_2^{T-1} & \mathcal{L_\text{mse}}_3^{T-1} & \cdots & \mathcal{L_\text{mse}}_{T-1}^{T-1} \\
  \mathcal{L_\text{mse}}_1^T & \mathcal{L_\text{mse}}_2^T & \mathcal{L_\text{mse}}_3^T & \cdots & \mathcal{L_\text{mse}}_{T-1}^T & \mathcal{L_\text{mse}}_T^T
\end{bmatrix}
\]
\vspace{-0.5\baselineskip}

The rows of the matrix represent evaluation scenarios, and the columns represent tasks. For example, $\mathcal{L_\text{mse}}_3^T$ denotes the test error on the third task after training on all $T$ tasks. 

Based on the matrix, this paper uses the following evaluation metrics. 

\vspace{-0.5\baselineskip}
\begin{equation*}
\label{eq:plasticity_loss}
\mathcal{L}_{\text{plasticity}} = \frac{1}{T}\sum_{t=1}^{T} \mathcal{L_\text{mse}}_t^t
\end{equation*}

\vspace{-0.5\baselineskip}
\begin{equation*}
\label{eq:stability_loss}
\mathcal{L}_{\text{stability}} = \frac{1}{T}\sum_{t=1}^{T} \mathcal{L_\text{mse}}_t^T
\end{equation*}

\vspace{-0.5\baselineskip}
\begin{equation*}
\label{eq:forgetting}
\mathcal{F} = \frac{1}{T}\sum_{t=1}^{T} \left( \mathcal{L_\text{mse}}_t^T - \mathcal{L_\text{mse}}_t^t \right)
\end{equation*}

\vspace{-0.5\baselineskip}
\begin{equation*}
\label{eq:forgetting-ratio}
\mathcal{F}_{\text{r}} = \frac{1}{T} \sum_{t=1}^{T} \frac{\text{max}(0, \mathcal{L_\text{mse}}_t^T - \mathcal{L_\text{mse}}_t^t)}{\mathcal{L_\text{mse}}_t^t}
\end{equation*}
\vspace{-0.5\baselineskip}

Lower $\mathcal{L}_{\text{plasticity}}$ and $\mathcal{L}_{\text{stability}}$ values indicate better estimation accuracy. $\mathcal{L}_{\text{plasticity}}$ is the average of the diagonal test errors in the above matrix. Each diagonal entry ($\mathcal{L_\text{mse}}_t^t$) represents the test error on task $t$ immediately after the model first learns that task. 
This paper uses $\mathcal{L}_{\text{plasticity}}$ to measure learning plasticity (i.e., the ability to rapidly adapt to new tasks).

$\mathcal{L}_{\text{stability}}$ is the average test error across all tasks in the last ($T$-th) scenario. This paper uses $\mathcal{L}_{\text{stability}}$ as a metric of memory stability (i.e., the ability to preserve past knowledge) because it measures how well the model retains performance on all learned tasks after continual training.

$\mathcal{F}$ measures the average change in task-wise test error between the time each task is first learned and the end of continual training. Higher $\mathcal{F}$ indicates greater forgetting, whereas lower $\mathcal{F}$ indicates better retention of past knowledge.  

$\mathcal{F}_{\text{r}}$  measures the average relative increase in test error from the time a task is first learned to the end of training. 
$\mathcal{F}_{\text{r}}~=~1$ means that the final error is twice the initial error (100\% relative increase). When $\mathcal{F}_{\text{r}} = 0$, no forgetting occurs.

\subsection{Results}

Table~\ref{tab:main-avgmse-stddev} summarizes the RTT estimation performance of AMR and the other methods in terms of $\mathcal{L}_{\text{plasticity}}$, $\mathcal{L}_{\text{stability}}$ and $\mathcal{F}$. Each result is averaged over 50 independent test runs. Boldface indicates the best result among the four methods. The * and ! symbols indicate that AMR's result is statistically different from the compared method under a $t$-test and Mann--Whitney $U$-test at $p < 0.05$, respectively. Standard deviations are reported with the $\pm$ symbol.

\vspace{-0.5\baselineskip}
\begin{table}[h]
\centering
\caption{RTT Estimation Performance}
\vspace{-0.5\baselineskip}
    \begin{tabular}{|l||c|c|c|}
    \hline
    \textbf{Method} & \textbf{$\mathcal{L}_{\text{plasticity}}$} & \textbf{$\mathcal{L}_{\text{stability}}$} & \textbf{$\mathcal{F}$} \\
    \hline
    \hline
    AMR & 0.1069$\pm$\textbf{0.0112} & \textbf{0.1399}$\pm$\textbf{0.0101} & \textbf{0.0330}$\pm$\textbf{0.0138} \\
    ACLT & 0.1073$\pm$0.0116 & 0.1457$\pm$0.0157 \textbf{*!} & 0.0384$\pm$0.0182 \\
    A2ER & \textbf{0.1057}$\pm$0.0127 & 0.1463$\pm$0.0158 \textbf{*!} & 0.0406$\pm$0.0173 \textbf{*!} \\
    CLeaR & 0.1075$\pm$0.0115 & 0.1454$\pm$0.0131 \textbf{*!} & 0.0379$\pm$0.0152 \\
    \hline
    \end{tabular}
\label{tab:main-avgmse-stddev}
\end{table}
\vspace{-0.5\baselineskip}

As shown in Table~\ref{tab:main-avgmse-stddev}, AMR achieves the lowest $\mathcal{L}_{\text{stability}}$ among the four methods, along with the lowest standard deviation. AMR also statistically outperforms the other three methods in $\mathcal{L}_{\text{stability}}$ under both the $t$-test and Mann--Whitney $U$-test ($p < 0.05$). For $\mathcal{F}$, AMR achieves the lowest value and the lowest standard deviation. These results indicate that AMR better preserves knowledge from previously learned tasks and maintains estimation accuracy across tasks. 

In $\mathcal{L}_{\text{plasticity}}$, A2ER achieves the lowest value, while AMR yields the second-lowest value and the lowest standard deviation. The difference between AMR and A2ER is small, suggesting that AMR maintains competitive plasticity while substantially improving stability. This indicates that AMR does not sacrifice its ability to adapt to newly encountered tasks in order to reduce forgetting.

\begin{figure}[t]
    \centering
    \resizebox{0.9\linewidth}{!}{\includegraphics{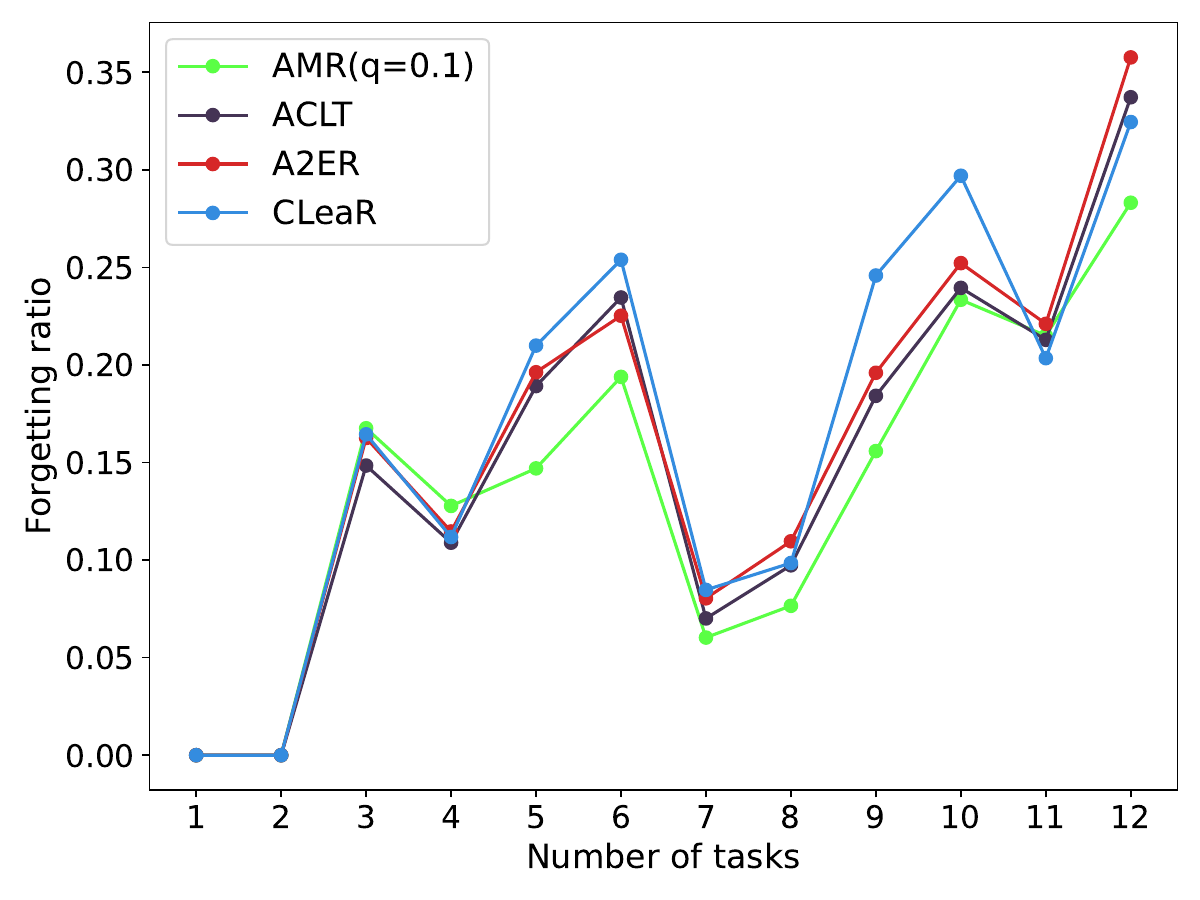}}
    \vspace{-0.5\baselineskip}
    \caption{Change of forgetting ratio as the number of sequential tasks varies}
    \label{fig:FR_main}
\end{figure}

Fig.~\ref{fig:FR_main} shows how the forgetting ratio ($\mathcal{F}_{\text{r}}$) changes as the number of sequential tasks increases. Overall, AMR attains the lowest average $\mathcal{F}_{\text{r}}$ at 0.1384, while ACLT, A2ER and CLeaR yield higher averages of 0.1518, 0.1596 and 0.1662, respectively. When the methods are ranked by $\mathcal{F}_{\text{r}}$ in each scenario, AMR achieves the best average rank of 1.9, while ACLT, A2ER and CLeaR have rank averages of 2.1, 3.0 and 3.0, respectively. 

Table~\ref{tab:main-times} shows the computational cost of training for AMR and the other  methods. The total training time refers to the time required to train on all tasks in the final evaluation scenario. CLeaR achieves the lowest training time, followed by A2ER and AMR. 
Although AMR requires slightly more training time than CLeaR and A2ER, its cost remains comparable to those methods and is lower than that of ACLT.

\begin{figure}[t]
    \centering
    \resizebox{0.8\linewidth}{!}{\includegraphics{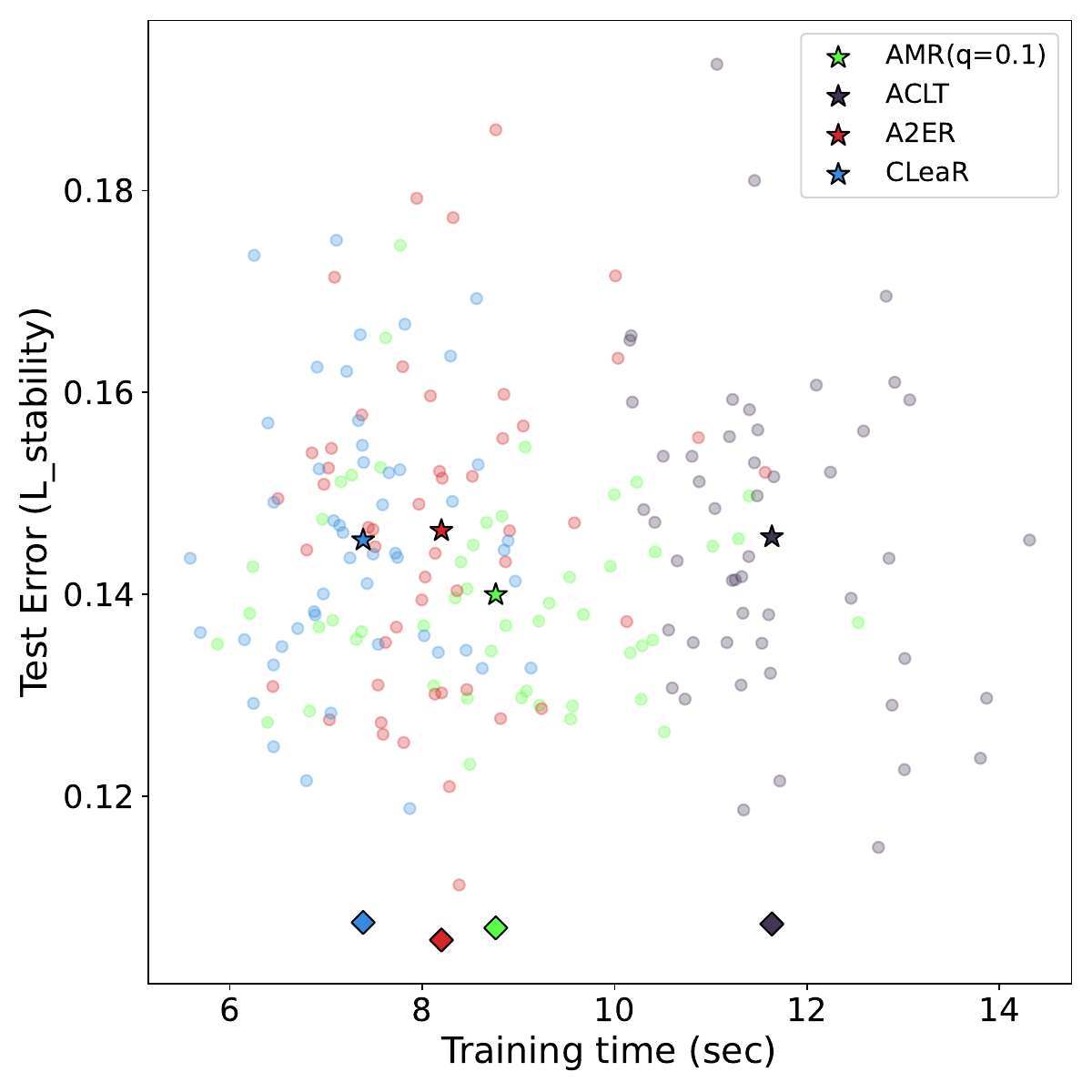}}
    \vspace{-0.5\baselineskip}
    \caption{Tradeoff between Estimation Performance and Training Cost}
    \label{fig:error_time_main}
\end{figure}

Fig.~\ref{fig:error_time_main} illustrates the tradeoff between estimation performance and training cost. Circular dots represent pairs of performance ($\mathcal{L}_{\text{stability}}$) and training time over 50 test runs, and their mean is plotted as a star. Diamonds represent the mean pairs of $\mathcal{L}_{\text{plasticity}}$ and training time. AMR achieves the lowest mean $\mathcal{L}_{\text{stability}}$ while maintaining a moderate training cost, indicating that it effectively maintains estimation performance on all previously seen tasks, not only on the most recent one, without imposing a substantial computational overhead.

\begin{table}[htb]
\centering
\caption{Computational cost for training}    
\vspace{-0.5\baselineskip}
    \begin{tabular}{|l||c|c|}
    \hline
    \textbf{Method} & \textbf{Total Training Time (sec)} & \textbf{Per-task Time (sec)} \\
    \hline
    \hline
    AMR & 8.76 $\pm$ 1.50 & 0.73 \\
    ACLT & 11.64 $\pm$ 1.02 & 0.97 \\
    A2ER & 8.20 $\pm$ 1.07 & 0.68 \\
    CLeaR & \textbf{7.39} $\pm$ \textbf{0.86} & \textbf{0.62} \\
    \hline
    \end{tabular}
\label{tab:main-times}
\end{table}

\vspace{-0.5\baselineskip}

Table~\ref{tab:amr-avgmse-stddev} summarizes how the synthesis range parameter $q$ (Algorithm~\ref{alg:amr}) affects the RTT estimation performance of AMR. The results show that AMR is relatively robust to the choice of $q$, with similar $\mathcal{L}_{\text{stability}}$ values across all tested settings. Among them, $q=0.1$ achieves the lowest $\mathcal{L}_{\text{stability}}$, while $q=0$ achieves the lowest $\mathcal{L}_{\text{plasticity}}$. This  suggests that interpolation-only synthesis can slightly improve adaptation to newly encountered tasks. The lowest $\mathcal{F}$ is obtained at $q=0.2$, indicating that a modest amount of extrapolation can further reduce forgetting. 

\begin{table}[t]
\centering
\setlength{\tabcolsep}{1.5pt}
\caption{RTT Estimation Performance (AMR)}
    \begin{tabular}{|c||c|c|c|}
    \hline
    \textbf{$q$} & \textbf{$\mathcal{L}_{\text{plasticity}}$} & \textbf{$\mathcal{L}_{\text{stability}}$} & \textbf{$\mathcal{F}$} \\
    \hline
    \hline
    0.0 & \textbf{0.1053} $\pm$ \textbf{0.0090} & 0.1408 $\pm$ 0.0120 & 0.0355 $\pm$ 0.0155 \\
    0.05 & 0.1079 $\pm$ 0.0105 & 0.1415 $\pm$ \textbf{0.0080} & 0.0335 $\pm$ \textbf{0.0123} \\
    0.1 & 0.1069 $\pm$ 0.0113 & \textbf{0.1399} $\pm$ 0.0101 & 0.0330 $\pm$ 0.0138 \\
    0.2 & 0.1097 $\pm$ 0.0106 & 0.1405 $\pm$ 0.0105 & \textbf{0.0308} $\pm$ 0.0128 \\
    0.3 & 0.1076 $\pm$ 0.0115 & 0.1412 $\pm$ 0.0120 & 0.0336 $\pm$ 0.0162 \\
    1.5 & 0.1060 $\pm$ 0.0124 & 0.1416 $\pm$ 0.0126 & 0.0356 $\pm$ 0.0150 \\
    2.5 & 0.1082 $\pm$ 0.0112 & 0.1414 $\pm$ 0.0125 & 0.0332 $\pm$ 0.0137 \\
    \hline
    \end{tabular}
\label{tab:amr-avgmse-stddev}
\end{table}

\begin{figure}[t]
    \centering
    \resizebox{0.9\linewidth}{!}{\includegraphics{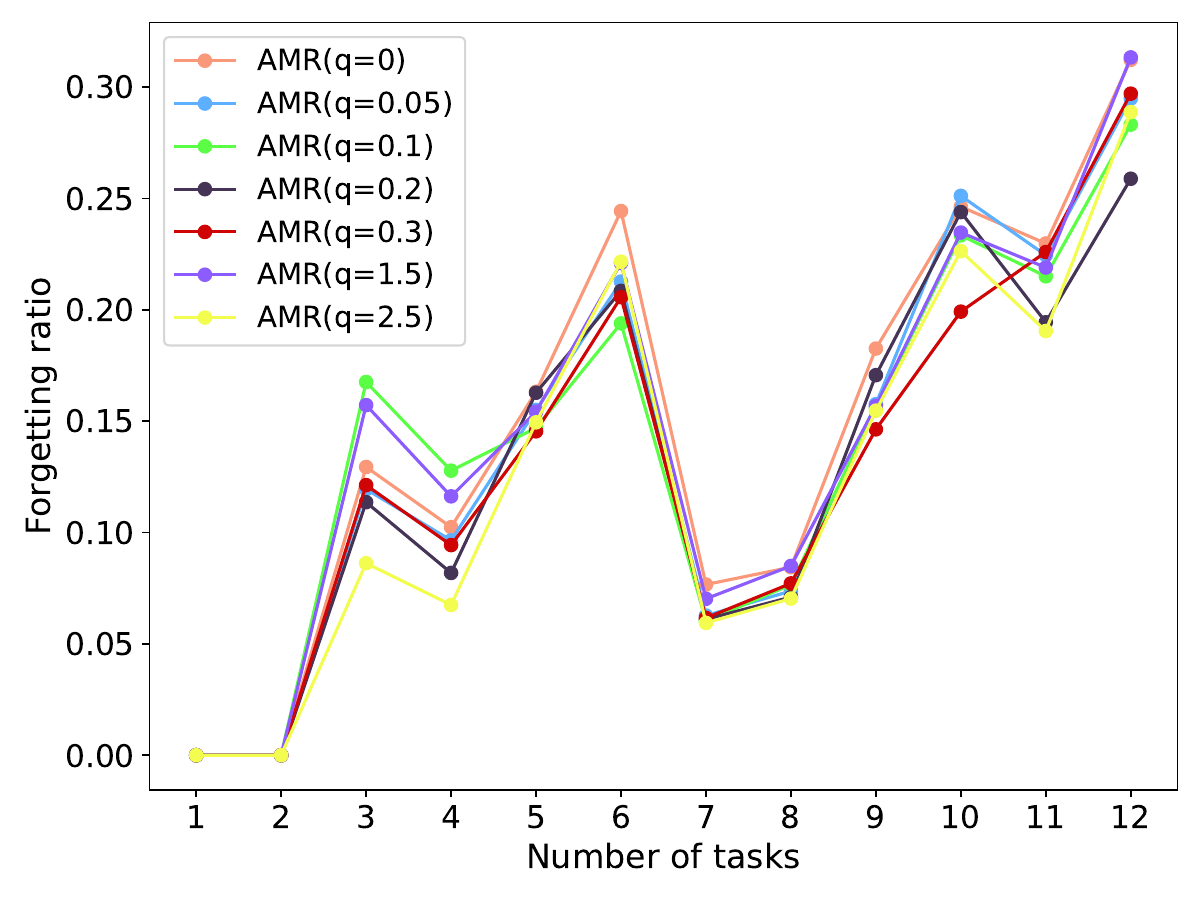}}
    \caption{Change in forgetting ratio as the number of tasks varies (AMR)}
    \label{fig:FR_amr_variations}
\end{figure}

Fig.~\ref{fig:FR_amr_variations} shows how $\mathcal{F}_{\text{r}}$ changes across sequential tasks for different $q$ values. The curves follow broadly similar trends, indicating that AMR's forgetting behavior is not highly sensitive to $q$. However, moderate values such as $q=0.1$ and $q=0.2$ tend to provide favorable forgetting mitigation, whereas very large extrapolation ranges do not yield clear additional benefits. This suggests that limited extrapolation is useful, but overly broad extrapolation may introduce synthetic samples that are less representative of previously learned tasks.

Table~\ref{tab:amr-times} shows the training cost of AMR under different $q$ values. The total training times are broadly comparable across settings, ranging from 8.76 to 10.29 seconds. $q=0.1$ achieves the lowest total training time, followed closely by $q=0.2$. This indicates that the choice of $q$ does not introduce substantial computational overhead, while moderate values of $q$ can improve stability or forgetting performance.

\begin{table}[htb]
\centering
\caption{Computational Cost for Training (AMR)}
    \begin{tabular}{|c||c|c|} 
    \hline 
    \textbf{$q$} & \textbf{Total Training Time (sec)} & \textbf{Per-task Time (sec)} \\ 
    \hline 
    \hline 
    0.0 & 9.46 $\pm$ 1.73 & 0.79 \\
    0.05 & 10.29 $\pm$ 1.99 & 0.86 \\ 
    0.1 & \textbf{8.76} $\pm$ 1.50 & \textbf{0.73} \\ 
    0.2 & 8.80 $\pm$ 1.51 & 0.73 \\ 
    0.3 & 9.10 $\pm$ 1.71 & 0.76 \\
    1.5 & 10.04 $\pm$ \textbf{1.19} & 0.84 \\ 
    2.5 & 9.14 $\pm$ 1.69 & 0.76 \\ 
    \hline 
    \end{tabular} 
\label{tab:amr-times} 
\end{table}

\begin{figure}[t]
    \centering
    \resizebox{0.8\linewidth}{!}{\includegraphics{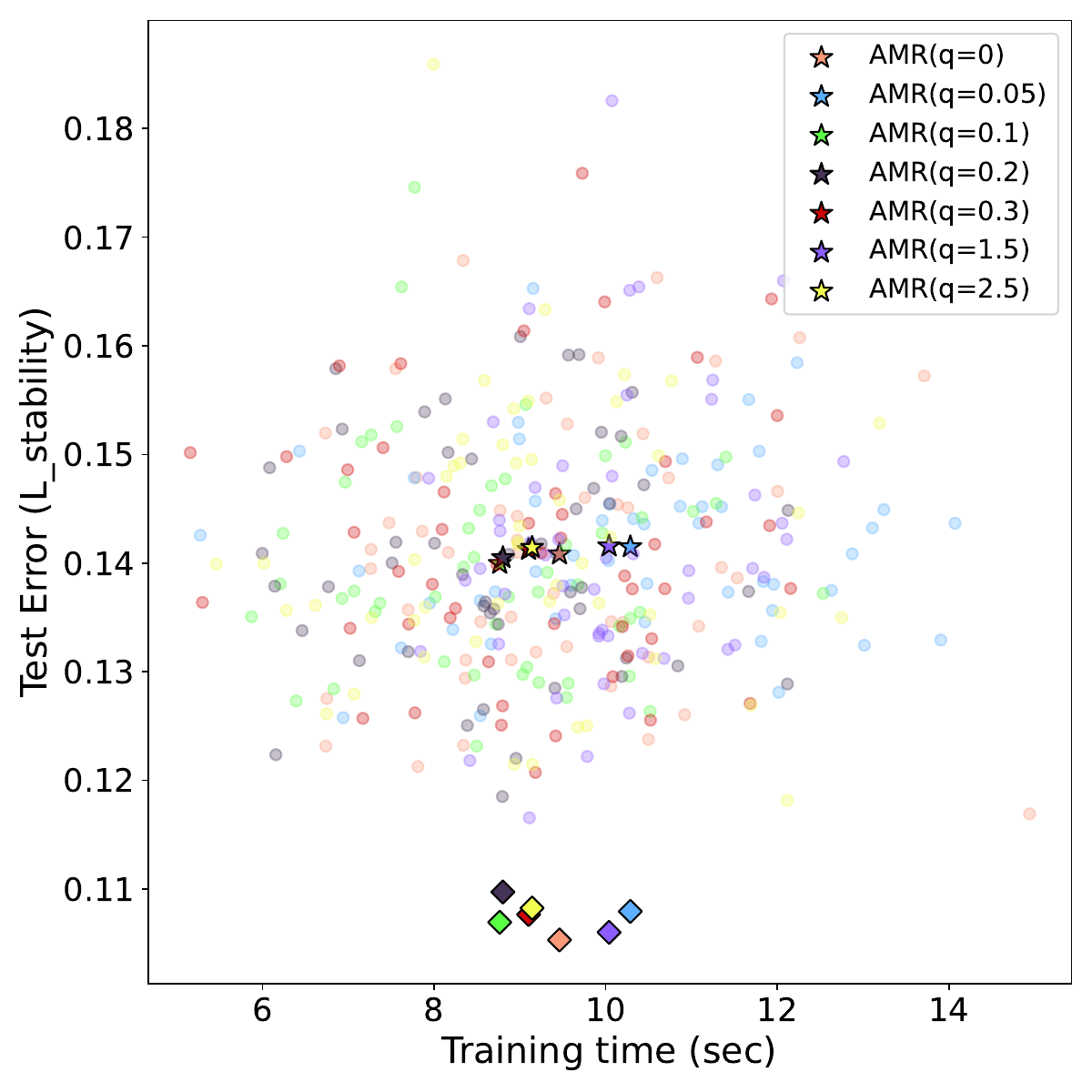}}
    \caption{Tradeoff between Estimation Performance and Training Cost (AMR)}
    \label{fig:error_time_amr_variations}
\end{figure}

Fig.~\ref{fig:error_time_amr_variations} illustrates the tradeoff between estimation performance and training cost for different $q$ values. The $\mathcal{L}_{\text{stability}}$ values are close across all cases. Overall, $q=0.1$ provides the best cost-performance tradeoff: it achieves the lowest $\mathcal{L}_{\text{stability}}$ and the lowest training time, while maintaining a low $\mathcal{F}$. Therefore, this paper uses $q=0.1$ as the default AMR configuration (Table~\ref{tab:train-hyper-params}). 

\section{Conclusion}

This paper proposes and evaluates Affine Memory Replay~(AMR), a continual learning (CL) method for RTT estimation in in-sequence molecular communication. AMR synthesizes replay samples from past simulation results through interpolation and extrapolation, stores both past and synthetic samples in bounded replay buffers, and uses replay losses to regularize model updates. Experimental results show that AMR significantly outperforms existing CL methods in   
maintaining overall estimation accuracy across sequentially learned estimation tasks. It effectively balances memory stability and learning plasticity under moderate training cost. Future work will extend AMR to more advanced molecular communication settings, including mobile bio-nanomachines, time-varying environments and larger task sequences. It will also investigate adaptive strategies for selecting the synthesis range parameter $q$ and broader replay-sample generation methods to further improve the balance between stability, plasticity and computational cost.

\bibliographystyle{IEEEtran}
\bibliography{ref}

\end{document}